\documentclass[runningheads]{llncs}
\usepackage{epsf}
\usepackage{multirow}
\usepackage{amssymb}
\usepackage{graphicx}

\usepackage{url}

\begin{document}

\title{Audio Summarization with Audio Features and Probability Distribution Divergence}
\titlerunning{Audio Summarization with Audio Features and P.D.D.}

\author{Carlos-Emiliano Gonz\'alez-Gallardo\inst{1} \and Romain Deveaud\inst{1} \and Eric SanJuan\inst{1} \and
Juan-Manuel Torres-Moreno\inst{1,2}}
\authorrunning{C.E. Gonz\'alez-Gallardo et al.}
%
\institute{LIA - Avignon Universit\'e , 339 chemin des Meinajaries, 84140, Avignon, France\\
\email{
\{carlos-emiliano.gonzalez-gallardo, eric.sanjuan, juan-manuel.torres\}@univ-avignon.fr\\
romain.deveaud@gmail.com} \and
D\'epartement de GIGL,  Polytechnique Montr\'eal,\\
C.P. 6079, succ. Centre-ville, Montr\'eal (Qu\'ebec) H3C 3A7 Canada}

\maketitle

\begin{abstract}

The automatic summarization of multimedia sources is an important task that facilitates the understanding of an individual by condensing the source while maintaining relevant information.
In this paper we focus on audio summarization based on audio features and the probability of distribution divergence.
Our method, based on an extractive summarization approach, aims to select the most relevant segments until a time threshold is reached.
It takes into account the segment's length, position and informativeness value.
Informativeness of each segment is obtained by mapping a set of audio features issued from its Mel-frequency Cepstral Coefficients and their corresponding Jensen-Shannon divergence score. 
Results over a multi-evaluator scheme shows that our approach provides understandable and informative summaries.


\end{abstract}

\section{Introduction}
\label{sec:Introduction}

Multimedia summarization has become a major need since Internet platforms like Youtube\footnote{https://www.youtube.com/} provide easy access to massive online resources.
In general, automatic summarization intends to produce an abridged and informative version of its source \cite{torres2014automatic}.
The type of automatic summarization we focus in this article is audio summarization, which source corresponds to an audio signal.

Audio summarization can be performed with the following three approaches: directing the summary using only audio features \cite{duxans2009audio,maskey2005comparing,maskey2006summarizing,zlatintsi2012saliency}, extracting the text inside the audio signal and directing the summarization process using textual methods \cite{christensen2008cascaded,rott_2016_Speechtotext,Taskiran_2006_Automated} and an hybrid approach which consists of a mixture of the first two \cite{szaszak2016summarization,zechner2003spoken,zlatintsi2015audio}. 
Each approach has advantages and disadvantages with regard to the others.
Using only audio features for creating a summary has the advantage of being totally transcript independent; however, this may also be a problem given that the summary is based only on how things are said.
By contrast, directing the summary with textual methods benefits from the information contained within the text, dealing to more informative summaries; nevertheless, in some cases transcripts are not unavailable. Finally, using both audio features and textual methods can boost the summary quality; yet, disadvantages of both approaches are present.

The method we propose in this paper consists of an hybrid approach during training phase while text independent during summary creation.
It resides on using textual information to learn an informativeness representation based on probability distribution divergences that standard audio summarization with audio features does not consider.
During the summarization process this representation is used to obtain an informativeness score without a textual representation of the audio signal to summarize.
To our knowledge, probability distribution divergences have not been used for audio summarization.

The rest of this article is organized as follows.
In Section \ref{sec:audio_sum} we give an overview of what audio summarization is, we include its advantages and disadvantages comparing it with other summarization techniques.
During Section \ref{sec:pdd_audio_sum} we explain how the probability distribution divergence may be used over an audio summarization framework and we describe in detail our summarization proposal. 
In Section \ref{sec:exp_set} we describe the dataset used during training and the summary generation phases as well as the evaluation metric that we adopted to measure the quality of the produced summaries and the results from the experimental evaluation of the proposed method.
Finally, Section \ref{sec:conclusions} concludes the article.

\section{Audio Summarization}
\label{sec:audio_sum}

Audio summarization without any textual representation aims to produce an abridged and informative version of an audio source using only the information contained in the audio signal. 
This kind of summarization is challenging because the available information corresponds to how things are said, this is advantageous in terms of transcripts availability.
Hybrid audio summarization methods or text based audio summarization algorithms need automatic or manual speech transcripts to select the pertinent segments and produce an informative summary \cite{zechner2003spoken,zlatintsi2015audio}.
Nevertheless, speech transcripts may be expensive, non available or of low quality, this creates repercussions over the summarization performance.

Duxans \textit{et al.} \cite{duxans2009audio} managed to generate audio based summaries of a soccer match using re-transmissions that detect highlighted events.
They based their detection algorithm on two acoustic features: the block energy and the acoustic repetition indexes.
The performance was measured in terms of goal recall and summary precision, showing high rates for both categories.

Maskey \textit{et al.} \cite{maskey2006summarizing} presented an audio based summarization method using a Hidden Markov Model (HMM) framework.
They used a set of different acoustic/prosodic features to represent the HMM observation vectors: speaking rate; F0 min, max, mean, range and slope; min, max and mean RMS energy; RMS slope and sentence duration.
The hidden variables represented the inclusion or exclusion of a segment within the summary.
They performed experiments over 20 CNN shows and 216 stories previously used in \cite{maskey2005comparing}.
Evaluation was made with standard Precision, Recall and F-measures information retrieval measures.
Results show us that the HMM framework had a very good coverage ($Recall = 0.95$) but a very poor precision ($Precision = 0.26$) when selecting pertinent segments.

Zlatintsi \textit{et al.} \cite{zlatintsi2012saliency} addressed the audio summarization task by exploring the potential of a modulation model for the detection of perceptually important audio events.
They performed a saliency computation of audio streams based on a set of saliency models and various linear, adaptive and nonlinear fusion schemes. 
Experiments were performed over audio data extracted from six 30-minute movie clips.
Results were reported in terms of frame-level precision scores showing that nonlinear fusion schemes perform best.

Audio summarization based only on acoustic features like fundamental frequencies, energy, volume change and speaker turn, has the big advantage that no textual information is needed.
This approach is especially useful when human transcripts are not available for the spoken documents and Automatic Speech Recognition (ASR) transcripts have a high word error rate.
However, for high informative contexts like broadcast news, bulletins or reports, most relevant information resides on the things that are said while audio features are limited to how things are said.

\section{Probability Distribution Divergence for Audio Summarization}
\label{sec:pdd_audio_sum}

\noindent All presented methods in the previous section omit the informativity content of the audio streams.
In order to overcome the lack of information, we propose an extractive audio summarization method capable of representing the informativeness of a segment in terms of its audio features during training phase; informativeness is mapped by a probability distribution divergence model.
Then, when creating a summary, textual independence is reached using only audio based features.

Divergence is defined by Manning \cite{Manning:1999:FSN:311445} as a function which estimates the difference between two probability distributions. 
In the framework of automatic text summarization evaluation, \cite{louis2008automatic,saggion2010Multilingual,Torres-Moreno10} have used divergence based measures such as Kullback–Leibler and Jensen–Shannon (JS) to compare the probability distribution of words between automatically produced summaries and their sources.
Extractive summarization based on the divergence of probability distributions has been discussed in \cite{louis2008automatic} and a method has been proposed in \cite{torres2014automatic} (DIVTEX).

Our proposal, based on an extractive summarization approach aims to select the most pertinent audio segments until a time threshold is reached. 
A training phase is in charge of learning a model that maps a set of 277 audio features to an informativeness value.
A big dataset is used to compute the informativeness by obtaining the divergence between the dataset documents and their corresponding segments.
During the summarization phase, the method takes into account the segment's length, position and the mapped informativeness of the audio features to rank the pertinence of each audio segment.

\subsection{Audio signal pre-processing}
\label{subsec:prepro}

During the pre-processing step, the audio signal is split into background and foreground channels.
This process is normally used on music records for separating vocals and other sporadic signals from accompanying instrumentation.
Rafii \textit{et. al} \cite{rafii2012music} achieved this separation for identifying recurrent elements by looking for similarities instead of periodicities.
 
Rafii \textit{et. al} approach is useful for those song records where repetitions happen intermittently or without a fixed period; however, we found that applying the same method to newscasts and reports audio files made much easier to segment them using only the background signal.
We assume this phenomena is due to the fact that newscasts and reports are heavily edited with a low volume of background music playing while the journalist speak and louder music/noises for transitions (foreground). 

Following \cite{rafii2012music}, to suppress non-repetitive deviations from the average spectrum and discard vocal elements, audio frames are compared using the cosine similarity.
Similar frames separated by at least two seconds are aggregated by taking their per-frequency median value to avoid being biased by local continuity.
Next, assuming that both signals are additive, a pointwise minimum between the obtained frames and the original signal is applied to obtain a raw background filter.
Then, a foreground and background time-frequency mask is derived from the raw background filter and the input signal with a soft mask operation.
Finally, foreground and background components are obtained by multiplying the time-frequency masks with the input signal.

\subsection{Informativeness model}
\label{subsec:infomo}

Informativeness is learned from the transcripts of a big audio dataset such as newscasts and reports.
A mapping between a set of 277 audio features and an informativeness value is learned during the training phase.
It corresponds to the Jensen-Shannon divergence ($D_{JS}$) between the segmented transcripts and their source.

The $D_{JS}$ is based on the Kullback-Leibler divergence \cite{kullback1951information} with the main difference that is symmetric. The $D_{JS}$ between a segment $Q$ and its source $P$ is defined by \cite{louis2009automatically,Torres-Moreno10} as:

\begin{eqnarray}
    D_{JS}(P||Q) &=& \frac{1}{2}\sum_{w \in P} \left [ P_w  \log_2 \left( \frac{2  P_w}{P_w + Q_w} \right) + Q_w  \log_2 \left( \frac{2  Q_w}{P_w + Q_w} \right) \right ] \\
    P_w &=& \frac{C^P_w + \delta}{|P| + \delta \times \beta} \\
    Q_w &=& \frac{C^Q_w + \delta}{|Q| + \delta \times \beta} 
\end{eqnarray}

where $C^{(P|Q)}_w$ is the frequency of word $w$ over $P$ or $Q$. To avoid shifting the probability mass to unseen events, the scaling parameter $\delta$ is set to $0.0005$. $|P|$ and $|Q|$ corresponds to the number of tokens on $P$ and $Q$.
Finally $\beta=1.5 \times |V|$, where $|V|$ is the vocabulary size on $P$.

Each segment $Q$ has a length of 10 seconds and is represented by 277 audio features where 275 corresponds to 11 statistical values of 25 Mel-frequency Cepstral Coefficients (MFCC) and the other two correspond to the number of frames in the segment and its starting time. The 11 statistical values can be seen in Table \ref{table:mfcc_features}, where $\phi'$ and $\phi''$ corresponds to the first and second MFCC derivative.

\begin{table}[!ht]
    \centering
    \begin{tabular}{lp{0.5cm}p{0.5cm}p{0.5cm}}
        \hline
        \textbf{Feature}& \multicolumn{3}{c}{\textbf{MFCC}}\\
        & $\phi$ & $\phi'$ & $\phi''$\\
        \hline
        min & \textbullet & &\\
        max & \textbullet\\
        median & \textbullet\\
        mean & \textbullet & \textbullet & \textbullet\\
        variance & \textbullet & \textbullet & \textbullet\\
        skewness & \textbullet\\
        kurtosis & \textbullet\\
        \hline
    \end{tabular}

\caption{MFCC based statistical values}
\label{table:mfcc_features}
\end{table} 

\begin{figure}
    \centering
    \includegraphics[width=0.75\textwidth]{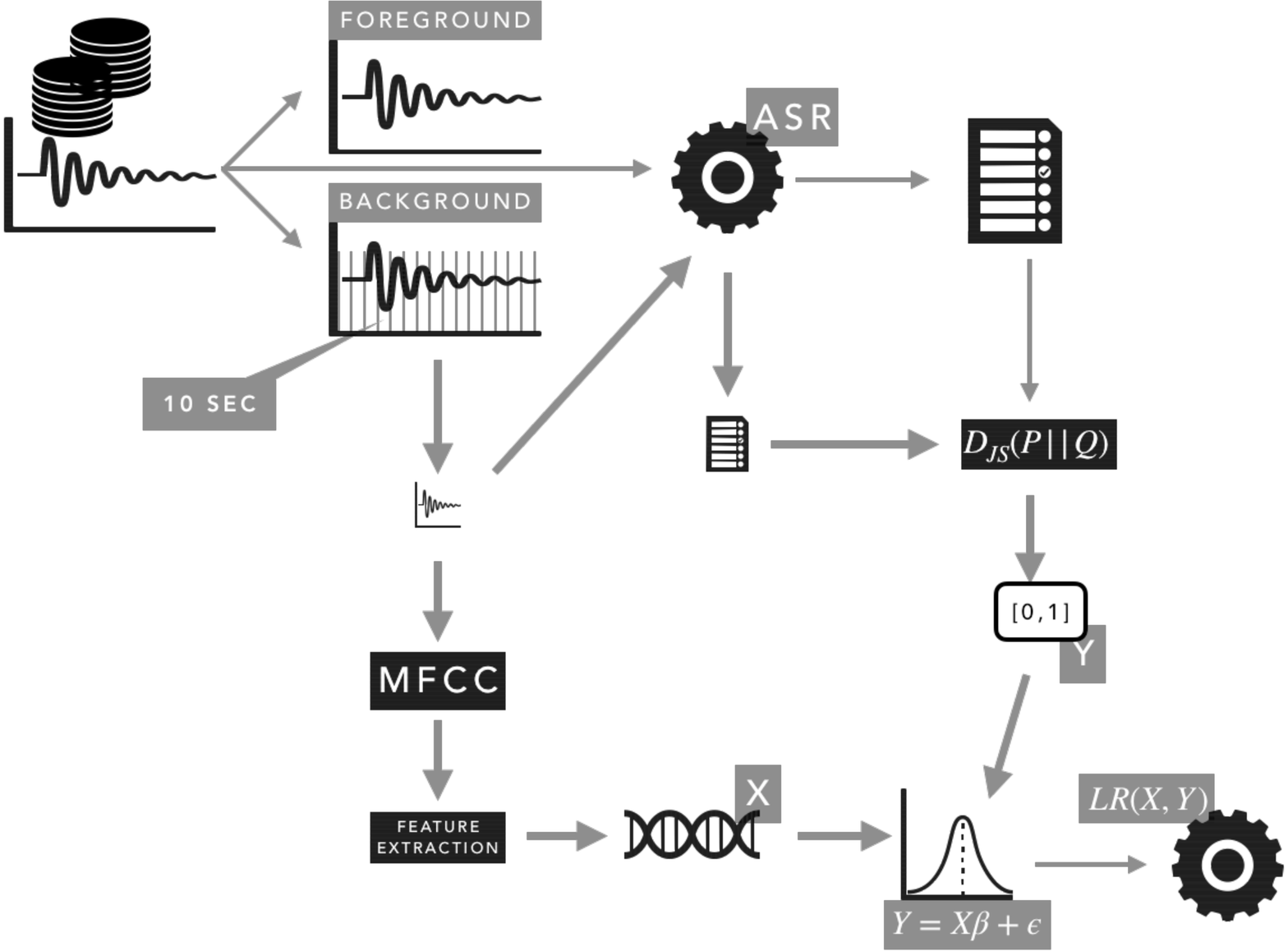}
    \caption{Informativeness model scheme}
    \label{fig:info_model}
\end{figure}

\noindent A linear least squares regression model ($LR(X,Y)$) is trained to map the 277 audio features ($X$) into a informativeness score ($Y$).
Figure \ref{fig:info_model} shows the whole training phase (informativeness model).
All audio processing and feature extraction is performed with the Librosa library \footnote{https://librosa.github.io/librosa/index.html} \cite{mcfee2015librosa}.

\subsection{Audio Summary Creation}

The summary creation of a document $P$ follows the same audio signal pre-processing steps described in Section \ref{subsec:prepro}. 
During this phase, only the audio signal is needed and informativeness of each candidate segment $Q_i \in P$ is predicted with the $LR(Q_i,Y_{Q_i})$ model.
Figure \ref{fig:sum_creation} shows the full summarization pipeline to obtain a $\theta$ threshold length summary of an audio document $P$.

After the background signal is isolated from the main signal a temporally-constrained agglomerative clustering routine is used to partition the audio stream into $k$ contiguous segments.

\begin{eqnarray}
    k =   \frac{P_{length}}{60} \times 20
\end{eqnarray}

being $P_{length}$ the length in seconds of $P$.

To rank the pertinence of each segment $Q_1 ... Q_k$, a score $S_{Q_i}$ is computed.
Audio summarization is performed by choosing those segments which contain higher $S_{Q_i}$ scores in order of appearance until $\theta$ is reached. $S_{Q_i}$ is defined as:

\begin{figure}
    \centering
    \includegraphics[width=0.75\textwidth]{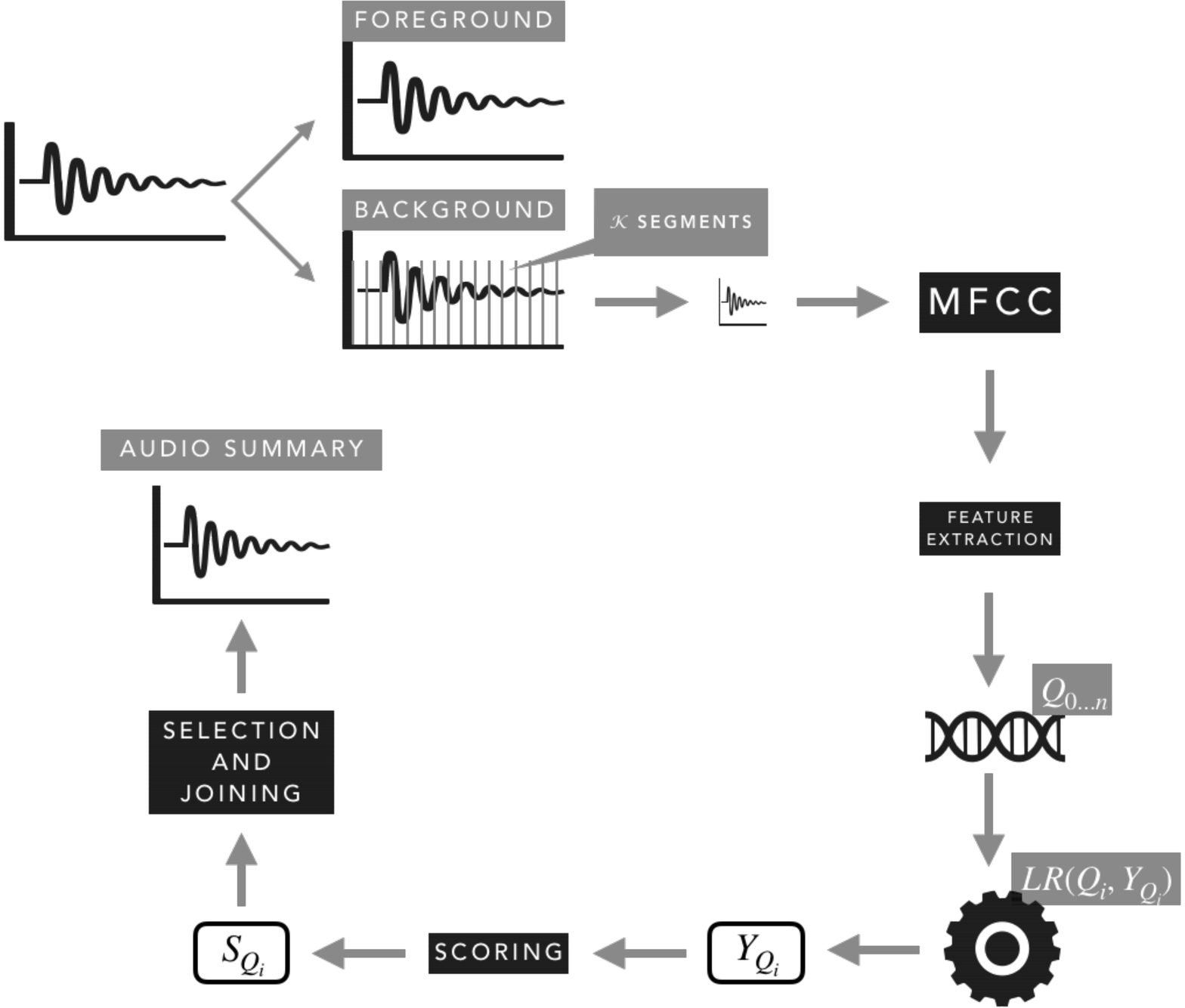}
    \caption{Summary creation scheme}
    \label{fig:sum_creation}
\end{figure}

\begin{eqnarray}
    S_{Q_i} &=& \frac{1}{1+e^{-(\Delta_{t_i} - 5)}} \times \frac{|Q_i|}{|P|} \times e^{- \frac{t_{Q_i}}{\Delta_{t_i}}} \times  e^{1-LR_{Q_i}}
\end{eqnarray}

Here $\Delta_{t_i} = t_{Q_i+1} - t_{Q_i}$, being $t_{Q_i}$ the starting time of the segment $Q_i$ and $t_{Q_i+1}$ the starting time of $Q_{i+1}$.
$|Q_i|$ and $|P|$ corresponds to the length in seconds of the segment $Q_i$ and $P$ respectively.

\section{Experimental Evaluation}
\label{sec:exp_set}

We trained the informativeness model explained in Section \ref{subsec:infomo} with a set of 5,989 audio broadcasts which corresponds to more than 310 hours of audio in French, English and Arabic \cite{miko2017video}.
Transcripts were obtained with the ASR system described on \cite{jouvet2018adaptation}.

During audio summary creation we focused on a small dataset of 10 English audio samples. In this phase no ASR system was used given the text independence our systems achieves once the informativeness model has been obtained.
Selected sample lengths vary between 102 seconds (1m42s) and 584 seconds (9m44s) with an average length of 318 seconds (5m18s).

Similar to Rott \textit{et al.} \cite{rott_2016_Speechtotext}, we implement a 1-5 subjective scaled opinion metric to evaluate the quality of the generated summaries and their parts. 
During evaluation, we provided a set of five evaluators with the original audio, the generated summary, their corresponding segments and the scale shown in Table~\ref{table:eval_scale}.

\subsection{Results}
\label{subsec:results}

Summary length was set to be the 35\% of the original audio length during experimentation.
Evaluation was performed over the complete audio summaries as well as over each summary segment.
We are interested on measuring the informativeness of the generated summaries but also on measuring the informativeness of each one of its segments.

\begin{table}[!ht]
    \centering
    \begin{tabular}{cl}
        \hline
        \textbf{Score}& \textbf{Explanation}\\
        \hline
        $5$ & Full informative\\
        $4$ & Mostly informative \\
        $3$ & Half informative\\
        $2$ & Quite informative\\
        $1$ & Not informative\\
        \hline
    \end{tabular}

\caption{Evaluation scale}
\label{table:eval_scale}
\end{table} 

Table \ref{table:audiosum_segscore} shows the length of each video and the number of segments that were selected during the summarization process.
``Full Score'' corresponds to the complete audio summaries evaluation while ``Average Score'' to the score of their corresponding summary segments.
Both metrics represent different things and seem to be quite correlated. ``Full Score'' quantifies the informativeness of all the summary as a whole while ``Average Score'' represents the summary quality in terms of the information of each of its segments.
To validate this observation, we computed the linear correlation between these two metrics obtaining a \textit{PCC} value equal to $0.53$.

\begin{figure}
    \centering
    \includegraphics[width=0.75\textwidth]{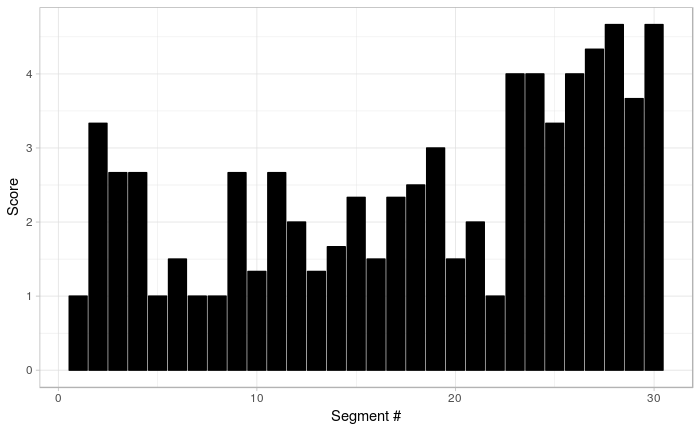}
    \caption{Audio summarization performance for sample \#6}
    \label{fig:performance_6}
\end{figure}

The average scores of all evaluators can be seen in Table \ref{table:audiosum_segscore}.
The lowest ``Full Score'' average value obtained during evaluation was $2.75$ and the highest $4.67$, meaning that the summarization algorithm generated at least half informative summaries.
``Average Score'' values oscillate between $2.49$ and $3.76$.
An interesting case is sample \#6, which according to its ``Full Score'' is ``mostly informative'' (Table \ref{table:eval_scale}) but has the lowest ``Average Score'' of all samples.
This difference is given because 67\% of its summary segments has an informativity score $<3$, but in general it achieves to communicate almost all the relevant information. Figure~\ref{fig:performance_6} plots the average score of each one of the 30 segments for sample \#6.

\begin{table}
    \centering
    \begin{tabular}{ccccc}
        \hline
        \textbf{Sample}& \textbf{Length}& \textbf{Segments} & \textbf{Full Score} & \textbf{Average Score} \\
        \hline
        1&  3m19s & 8& 4.20& 2.90\\
        2&  5m21s & 13& 3.50& 2.78\\
        3&  2m47s & 5& 3.80& 3.76\\
        4&  1m42s & 5& 3.60& 2.95\\
        5&  8m47s & 22& 4.67& 3.68\\
        6&  9m45s & 30& 4.00& 2.49\\
        7&  5m23s & 8& 3.20& 3.75\\
        8&  6m24s & 20& 3.75& 2.84\\
        9&  7m35s & 18& 3.75& 3.19\\
        10& 2m01s &  4& 2.75& 2.63\\
        \hline
    \end{tabular}

\caption{Audio summarization performance over complete summaries and summary segments}
\label{table:audiosum_segscore}
\end{table}

\begin{figure}
    \centering
    \includegraphics[width=1.0\textwidth]{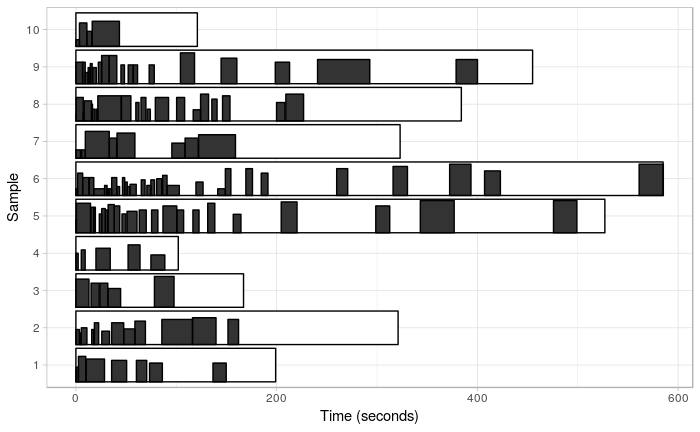}
    \caption{Graphical representation of audio summarization performance}
    \label{fig:audiosum_segscore_distrib}
\end{figure}

A graphical representation of the audio summaries and their performance can be seen in Figure \ref{fig:audiosum_segscore_distrib}. 
Full audio streams are represented by white bars while summary segments are represented by the gray zones.
The height of each summary segment corresponds to their informativeness score.

From Figure \ref{fig:audiosum_segscore_distrib} it can be seen that samples \#2, \#3, ,\#7, \#8 and \#10 have all their summary segments clustered to the left.
This is due to the preference that the summarization technique is given to the first part of the audio stream region whereby, within a standard newscast, is gathered the major part of the information.
The problem is that in cases where different topics are covered over the newscast (multi-topic newscast, interviews, round tables, reports, etc.), relevant information is distributed all over the video.
If a big amount of relevant segments are grouped in this region, the summarization algorithm uses all the space available for the summary very fast, discarding a large region of the audio stream.
This is the case of samples \#7 and \#10 which ``Full Scores'' are less to $3.50$.

Concerning sample \#5, a well distribution of its summary segments is observed.
From its 22 segments, only 4 had an informativeness score $\leq3$, achieving the highest ``Full Score'' of all samples and a good ``Average Score''.


\newpage

\section{Conclusions}
\label{sec:conclusions}

In this paper we presented an audio summarization method based on audio features and on the hypothesis that mapping the informativeness from a pre-trained model using only audio features may help to select those segments which are more pertinent for the summary.

Informativeness of each segment was obtained by mapping a set of audio features issued from its Mel-frequency Cepstral Coefficients and their corresponding Jensen-Shannon divergence score. 
Summarization was performed over a sample of English newscasts, demonstrating that the proposed method is able to generate at least half informative extractive summaries.
We can deduce that there is not a clear correlation between the quality of a summary and the quality of its parts.
However this behavior could be modeled as a recall based relation between both measures.

As future work we will validate this hypothesis as well as expand the evaluation dataset from a multilingual perspective to consider French an Arabic summarization.

\section*{Acknowledgments}

We would like to acknowledge the support of CHIST-ERA for funding this work through the Access Multilingual Information opinionS (AMIS), (France - Europe) project.


\bibliographystyle{splncs04}
\bibliography{paper}
\end{document}